\documentclass[11pt]{article}

\usepackage[T1]{fontenc}
\usepackage[utf8]{inputenc}

\usepackage[usenames,dvipsnames]{color}

\usepackage{lmodern}
\usepackage{hyperref}
\usepackage{graphicx}
\usepackage{pdfpages}
\usepackage{amsmath}
\usepackage{amssymb}
\usepackage{a4}
\usepackage{indentfirst}
\usepackage{fancyhdr}
\usepackage{varioref}
\usepackage{makeidx}

\usepackage{mslapa}

\usepackage[normalem]{ulem}

\title{\Huge \textbf{Predicting the consequence of action in digital control state spaces} }
\author{\textsc{Emmanuel Daucé}}

\begin{document}

\maketitle
\begin{abstract}

The objective  of this dissertation is to shed light on some fundamental impediments 
in learning control laws in continuous state spaces. 
In particular, if one wants to build artificial devices capable to learn motor tasks
the same way they learn to classify signals and images, 
one needs to establish control rules that \textit{do not necessitate} comparisons between quantities
of the surrounding space.
We propose, in that context, to
take inspiration from the ``end effector control'' principle, as suggested by neuroscience studies,  
as opposed to the ``displacement control'' principle used in the classical control theory. 

\end{abstract}

\section{Problem statement}

Most machine learning techniques rely on a pattern matching principle, with representative vectors constructed from many passes over large datasets of examples.
The resulting set of prototypes then works as a dictionary of shapes, providing a reduced description of the input data. The projection of the input vectors on this redescription space is then expected to facilitate further data processing and interpretation.
Learning then coincides with the construction of a vocabulary of shapes that serve as a key to interpret the data. 
There are however few examples in the literature where auto-encoding algorithms end up in learning a control law in a continuous actions space.  
This difficulty seems to stem from the conceptual separation between, on one side, the dynamical systems approach and, on the other side, the pattern matching principles.

From the dynamical system perspective, a usual simplification considers a controller and its environment as a single dynamical system, with a reciprocal coupling between an internal part devoted to data processing and control and an external part consisting in material objects. The transformations from the internal state space toward the external state space are done through actuators. Reciprocally, the transformations from the external state space toward the internal state space are done through captors . 
When  the question of learning and updating actions is considered, some inconsistencies may however rapidly show off.

\section{Quick mathematic overview}

\subsection{Closed-loop control}
Closed-loop control systems are generally described using two state spaces, namely the controller space $\mathcal{X}_\text{in}$, whose state at time $t$ is $\boldsymbol{x}^\text{in}_t \in\mathcal{X}_\text{in} $ and the physical space $\mathcal{X}_\text{out}$, whose state at time $t$ is $\boldsymbol{x}^\text{out}_t \in\mathcal{X}_\text{out}$. The general evolution of the system is provided by a set of 4 stochastic update equations, i.e.:
  \begin{align}
  &{\boldsymbol{x}}^\text{out}_{t+dt} \sim P_\text{out}(\boldsymbol{x}^\text{out}_t,\boldsymbol{u}_t) \label{eq:feed-noise-1} \\
  &\boldsymbol{I}_t \sim Q_\text{in}(\boldsymbol{x}^\text{out}_t) \label{eq:feed-noise-2}\\
  &{\boldsymbol{x}}^\text{in}_{t+dt} \sim P_\text{in}(\boldsymbol{x}^\text{in}_t,\boldsymbol{I}_t) \label{eq:feed-noise-3} \\
  &\boldsymbol{u}_t \sim Q_\text{out}(\boldsymbol{x}^\text{in}_t) \label{eq:feed-noise-4}
  \end{align}
where eq. (\ref{eq:feed-noise-1}) is the external update,  eq. (\ref{eq:feed-noise-2}) the sensors actualization, eq. (\ref{eq:feed-noise-3}) the internal update and  eq. (\ref{eq:feed-noise-4})  the motor realization, with $\boldsymbol{I}_ t$ the sensory input,  $\boldsymbol{u}_t$ the command, $P_\text{out}$ the (external) generative process, $Q_\text{in}$ the measure, $P_\text{in}$ the internal program, and  $Q_\text{out}$ the command transmission chain.

\subsection{Active sensing}

In most cases, the external process is considered hidden, and only the sensory consequence $\boldsymbol{I}_t$ of the hidden state is observable. Then, learning is mainly inferring the parameters $\Theta$ that describe both the (inferred) generative and observation processes, $\hat{P}$ and $\hat{Q}$, i.e. 
\begin{align}
\max_{\boldsymbol{x},\Theta} L(\boldsymbol{x},\Theta)
\end{align}
with  $L(\boldsymbol{x},\Theta)$ the likelihood of the observation (given parameters $\Theta$ and a past estimation on $\boldsymbol{x}_t$):
\begin{equation}
L(\boldsymbol{x},\Theta) = \hat{Q}(\boldsymbol{I}_{t+dt}|{\boldsymbol{x}},\Theta)\hat{P}({\boldsymbol{x}}|{\boldsymbol{x}}_t,\boldsymbol{u}_t,\Theta) \label{eq:lik}
\end{equation}

which is intractable in the general case and must be decomposed in a marginalization ``expectation'' step:
$$\hat{f}(\boldsymbol{x}) = L(\boldsymbol{x}|\Theta)$$
and a parameter maximization step \shortcite{Dempster1977}:
$$\max_{\Theta'} \int L(\boldsymbol{x},\Theta')
\hat{f}(\boldsymbol{x}) d\boldsymbol{x} \label{eq:Dempster}$$

The active inference paradigm, as proposed by \shortcite{Friston2009}, mainly innovate from this classical scheme by including the command in the optimization, i.e. 
\begin{align}
\max_{\boldsymbol{x},\boldsymbol{u},\Theta} L(\boldsymbol{x},\boldsymbol{u},\Theta)
\end{align}
which tells in short that the meaning of action is to participate in the likelihood maximization process by sampling regions of the sensory space that should reduce the uncertainty about the hidden state. The choice of action $\boldsymbol{u}_t$ is there related to the maximization of the likelihood of the future hidden state estimate $\boldsymbol{x}_{t+dt}$ in the next observation step (see eq.\ref{eq:lik}), with action outcome estimated from current $\hat{P}$ and $\hat{Q}$. A particular modeling effort needs thus to be put on estimating the sensory prediction of action (forward model of action), which is the essence of the predictive coding scheme.  
The approach, however, doesn't provide a insight on which motor control schemes are effectively suitable to implement such prediction.   

Classical motor control schemes consider for instance actions as changes taking place in the physical (Euclidian) space where objects are defined by specific 3D coordinates. Active sensing means thus moving the eyes, neck and trunk in order to estimate the nature and location of the objects that compose the scene.  
Many referentials may be defined in relation to the different degrees of freedom of the main articulations taking part in scanning the visual environment, with e.g. the egocentric referential (trunk referential), the head-centered referential (relative to the direction of the head) and the retinocentric referential (relative to the direction of the sight). Since relevant referentials and DOFs are defined, an action mainly reduces to a referential change on an invariant scene. For instance, moving the eyes to the right translates the visual field to the left in the direction opposite to the angular command. The  linear transformation associated with action has a predictable consequence on where the objects will project on it after change. 

\section{Learning to predict}

Realizing and implementing those predictive transformations in a neuronal circuit (suitable for learning) happens however to be more difficult than expected. This difficulty is mainly due to the absence of a definite (operational) distance in the control space that implements learning, on contrary to the physical space that implements action.  

When learning is considered, the universality of a learning scheme  (its capability to learn any transformation) mostly relies on using non-linear transfer functions at one step (at least) of the data processing pathway, with, e.g. threshold functions, argmax, sigmoid, kernels etc.  \shortcite{Hay99}. The many modern learning techniques used in pattern recognition rely on those non-linear steps to gain enough discriminative power and extract common features from many samples. The pattern matching approach thus does not guarantee the proportionality of treatments between close-by and distant parts of the physical space.   In other words, the distance that can be defined into the redescription space may not  respect the distance and topology of the initial space. The redescription space is generally considered as a ``qualitative'' space (or ``digital'' space), containing features,  by contrast with the external ``quantitative'' space. 

In any case, as soon as the initial distances between the physical objects are not preserved, predicting the consequence of an action may necessitate a specific training (instead of considering explicit geometric transformations). If every sensory modality of the physical space is distributed on many neurons (whose receptive field may only respond to a particular interval or ``pixel''), then simple transformations such as translations, rotations (and even identity) need to be learned ``by heart'' \cite{Pouget1997} i.e. may not result from local operations on graded firing rate inputs. 
Even feasible at arbitrary precision in theory, learning ``by heart'' the sensory consequences of action may thus suffers from a superlinear (squared) complexity when distances or referential transformations are considered. Learning, for instance, to predict the visual field consequence of visual orientation commands would necessitate to 
learn every combination of objects under the many possible eye and head orientation with trunk-head and head-eye referential transfers, which may expectedly be very costly \shortcite{Pouget1997,Pouget2002}.

\section{Alternate view}



If one expects build artificial devices that learn motor tasks the same way they may learn to classify images and signals, one needs to reconsider motor control from the basements, in a way that renders compatible the auto-encoding and the predictive coding principles.  
This first implies a digital description of motor commands and motor acts, i.e a decomposition of the task space in elementary components, the same way the sensory state space may de decomposed in elementary constituents. In counterpart, the isometric transformations and quantitative comparisons used in classical control theory should be eliminated when possible to avoid combinatorial explosion in learning ``by heart'' the linear (euclidian) transformations and distances in the controller state space. 

In order to do predictive coding without learning transformations and distances, what one needs to do is directly learn a mapping from the motor command space toward the sensory space. 
In complex bodies made of many joints, a motor space is made of many dimensions, and the spatial disposition of the different segments of the body (posture) are unambiguously defined by their respective coordinates. In particular, the orientation of the trunk, the neck and the sight (relatively to the ground) are unambiguously set. 

When visual prediction is considered, a simple mapping of the postural information toward the visual field is possible.
If, for simplicity, only the hand and vision are considered, many visual orientations may produce many predictable  visual fields for the same invariant hand position. Reversely, moving the hand or the fingers while keeping the eyes  fixed also has predictable visual consequences. 
The visual consequences of all possible combinations of eyes and hand positions may thus be learned in a straightforward way without intermediary transformation. 

This has important consequences on the way motor control should be considered. A traditional view considers motor acts as displacements that are proportional to a certain positional error detected by the senses. In that perspective, a \emph{difference} between the desired external state $\boldsymbol{x}^*$ and the actual estimate $\hat{\boldsymbol{x}}$ needs to be calculated to issue a displacement command $\boldsymbol{u}$. This is the \emph{error-based} control framework and the \emph{displacement-based} command. There is thus no direct mapping from a displacement  $\boldsymbol{u}$ toward its visual consequence $\boldsymbol{I}$. One needs to pass through an intermediary object, the ``external state'' $\hat{\boldsymbol{x}}$  to perform a prediction. 

An alternate view is to consider the \emph{end-effector}-based (or ``ballistic'') control framework. 
Direct stimulations of the motor cortex have shown converging movements toward a same final posture whatever the initial position  \shortcite{Graziano2002}. The presence of motor vector fields were also observed in many species, and vector-field-based control was theorized by e.g. \shortcite{Mussa2004}. The vector field approach to control considers the end-position of the limb as the relevant control level over which the brain operates. An end-position mainly consists in the distal spatial coordinates of certain effector (tentacle, tongue or arm in animals). A motor vector field is a definite set of flexion/extension commands whose combination mechanically conducts the effector toward a particular end-point. To control an end-effector in 2D coordinates for instance, a simple referential can be set with three fields (whose end-points are not colinear). Then, controlling the effector within the triangle made by the three end-points is done by linear combinations. A dot in the coordinate system is defined by three weights affected to the different fields. 
This principle is generalizable to 3D physical spaces (four dots at least) and of course to the more complex (many degrees-based) postural states like, e.g., the fingers of the hand.

An interesting consequence, when the arm is considered, is the correspondence of every motor command with a particular position of the hand (end-effector) in the peripheral space. 
One can thus postulate the hand-and-arm-controlling motor cortex to be organized like a map, where every particular position in the map codes for a specific ``motor primitive'' (a linear combination of motor fields) that may conduct the hand toward a particular position in the peripheral space.
This (approximate) correspondence between a  motor command and a corresponding 3D coordinate in peripheral space renders  possible a direct prediction from the command $\boldsymbol{u}_\text{arm}$ toward its visually expected consequence $\boldsymbol{I}$ (as far as the hand is considered).

Despite fewer biological evidence, the very same reasoning can be applied on eye orientation effectors.  Orienting the eye toward the peripheral space can be seen as setting the angular direction $\boldsymbol{u}_\text{eye}$ of the central part of the retina (fovea) in rotational coordinates.
As far as only the hand and the sight are considered, and ignoring all possible luminance, background and light incidence changes, a direct mapping from $(\boldsymbol{u}_\text{arm}, 
	\boldsymbol{u}_\text{eye})$ toward $\boldsymbol{I}_\text{visual field}$ can be considered for learning, from a systematic scan of every possible eye and arm command. This is the equivalent of the ``babbling'' seen in young humans and animals, also addressed in epigenetic robotics \shortcite{Andry2001}. 

A direct mapping and prediction from the action space toward the sensory space is thus made possible in a digital control system provided specific ballistic control laws are considered. This first and preliminary insight is not a definite answer to the addressed question, but may help to reorient and radically simplify the learning control framework in order to address more challenging questions. 

\section{Concluding remarks}

A first remark, as soon as  brain modeling is considered, is the probable combination of several control principles serving different purposes in biological systems. 
If end-effector control happen to have interesting predictive capabilities, a counterpart is a certain lack of adaptivity in changing conditions.  Changes in body tilt, muscle fatigue, or any effector flaw may have a dramatic consequence on the final precision of movement. A parallel and fast-adapting control is expected to take place in the cerebellum, with the capability to deviate the motor orders on the fly at the time it is sent to the muscles.  The cerebellum has been shown to be sensible to the motor error, and capable to calculate a displacement correction \shortcite{Wolpert1998,Haruno2001,Dean2010}. The huge number of neurons implied in that calculation may reflect the combinatorial problem encountered as soon as positional differences need to be calculated to set a command. 
 
A follow-up of the hand/eye end-effector learning perspective is the learning the manipulation of objects by hand. The many different poses of the different objects lying in the surrounding space are putatively linkable to the way they can be handled by the hand. This is the essence of object affordances postulated by Gibson \cite{Gib79}. Object affordances (when handling by hand) is here an essential part of their identity. Scanning an object identity means here trying different manipulation schemes and inferring their visual and proprioceptive predictible consequences. An object in space is thus, in short, assimilated with its corresponding grasping operation, including the hand aperture and orientation when catching the object. From this motor perspective, many visual viewpoints may associate unambiguously different visual fields with this single invariant motor viewpoint. 

If the position of objects may, in principle, be learned from this end-effector prospect, the detailed linkage and decomposition in features and components is still an open-ended question. For instance, as soon as hand manipulation is concerned, a detailed and exhaustive exploration of each and every eye orientation/end-effector configuration should reveal impossible in practice. Random-driven exploration may rather be considered, with
exploration policies that should give advantage to surprise (unexpected changes) to known expectations. 
The final outcome may thus be the modeling of the construction of objects, i.e. the assimilation (or generalization) of many perspectives in a single motor framework, as postulated by Piaget in the sensori-motor stage \cite{Piaget1973}. This generalization capability, from multiple views toward a single concept, may correspond to the effective interpretation of an external ``cause'' $\hat{\boldsymbol{x}}$ to the actual sensation.

\end{document}